%% file: main.tex
\pdfoutput=1

\documentclass[11pt]{article}

\usepackage{EMNLP2022}
\usepackage{times}
\usepackage{latexsym}
\usepackage[T1]{fontenc}
\usepackage{listings}
\usepackage[utf8]{inputenc}

\usepackage{microtype}

\usepackage{microtype}
\usepackage{graphicx}
\usepackage{subfigure}
\usepackage{booktabs} %
\usepackage[english]{babel}
\usepackage{amsmath}
\usepackage{amssymb}
\usepackage{bbm}
\usepackage{inconsolata}
\usepackage{booktabs}
\usepackage{setspace}
\usepackage{tabularx}

\usepackage{enumitem}
\setitemize{noitemsep,topsep=0pt,parsep=0pt,partopsep=0pt}
\usepackage{xspace}
\usepackage[capitalize,noabbrev]{cleveref}

\newcommand{  \stderr}[1]{\scriptsize $\pm #1$}

\newcommand\influencemethod{\textsc{TracIn}\xspace}
\newcommand\embeddingmethod{\textsc{Embed}\xspace}

\usepackage{pmboxdraw}

\usepackage{tabularx}
\usepackage{CJKutf8}

\definecolor{myorange}{RGB}{216,108,34}
\definecolor{myblue}{RGB}{40,96,143}
\definecolor{mygreen}{RGB}{48,131,44}

\title{Towards Tracing Factual Knowledge in Language Models Back to the Training Data}

\author{
Ekin Akyürek$^{\dagger}$\And
Tolga Bolukbasi \And
Frederick Liu \And
Binbin Xiong \AND
Ian Tenney \And
Jacob Andreas$^{\dagger}$ \And
Kelvin Guu \AND
\normalfont{Google Research} \quad \normalfont{$^{\dagger}$MIT CSAIL}
}

\begin{document}
\maketitle

\begin{abstract}
Language models (LMs) have been shown to memorize a great deal of factual knowledge contained in their training data. But when an LM generates an assertion, it is often difficult to determine \emph{where} it learned this information and whether it is true.
In this paper, we propose the problem of \emph{fact tracing}: identifying which training examples taught an LM to generate a particular factual assertion. Prior work on \emph{training data attribution} (TDA) may offer effective tools for identifying such examples, known as ``proponents''. We present the first quantitative benchmark to evaluate this.
We compare two popular families of TDA methods --- \emph{gradient}-based and \emph{embedding}-based --- and find that much headroom remains. For example, both methods have lower proponent-retrieval precision than an information retrieval baseline (BM25) that does not have access to the LM at all.
We identify key challenges that may be necessary for further improvement such as overcoming the problem of gradient saturation, and also show how several nuanced implementation details of existing neural TDA methods can significantly improve overall fact tracing performance. 
\footnote{Code for the experiments is released at \url{https://github.com/ekinakyurek/influence}, and the datasets can be downloaded from \url{https://huggingface.co/datasets/ekinakyurek/ftrace}. Correspondences to \href{mailto:akyurek@mit.edu}{akyurek@mit.edu}}
\end{abstract}
\section{Introduction}
\label{sec:intro}
\begin{figure}[t]
    \centering
\includegraphics[width=\linewidth]{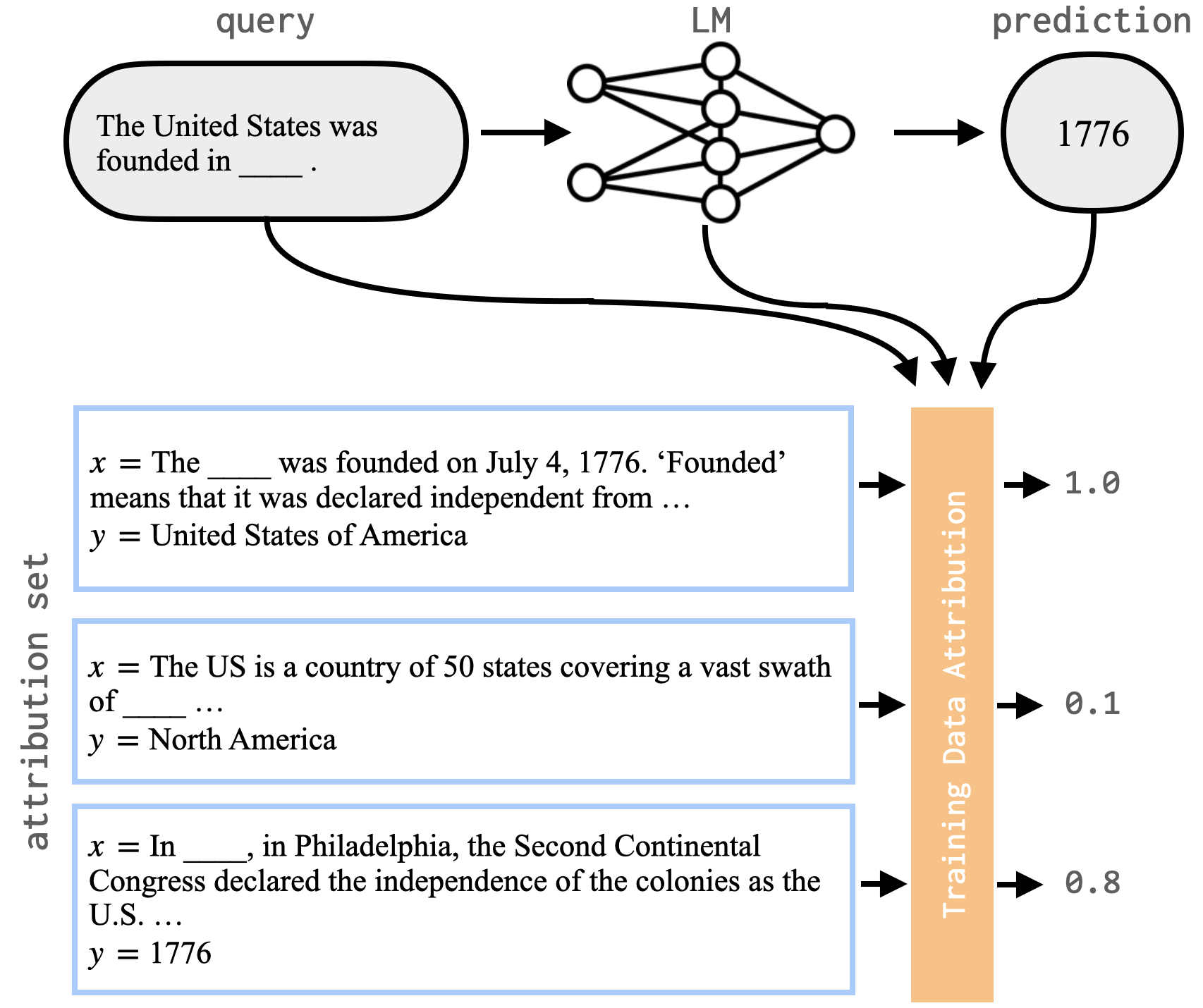}
    \caption{FTRACE benchmark for tracing a language model's predictions back to training examples (``proponents''): We provide two fact attribution datasets: one with real facts (\textsc{FTRACE-TREx}) and one with synthetic facts (\textsc{FTRACE-Synth}). We evaluate commonly studied attribution methods, including gradient-based and embedding-based approaches for their ability to identify true proponents.\vspace{-1em}}
    \label{fig:teaser}
\end{figure}
Research has shown that language models (LMs) acquire significant amounts of world knowledge from the massive text corpora on which they are trained \cite{petroni2019language, raffel2019exploring}. This development has enabled exciting advances in knowledge-intensive NLP tasks such as open-domain question answering \cite{roberts2020much} and knowledge base population \cite{petroni2019language}.
LMs have also been shown to generate factually incorrect statements \cite{lee2019hallucinations, tian2019sticking}, which is problematic for many applications where trustworthiness is important. Hence, there is an urgent need to understand exactly how LMs acquire and store knowledge so that we may improve their accuracy and coverage.
\paragraph{Training Data Attribution}
Ultimately, a language model's ``knowledge'' must derive from its training data.
But there has been little research on attributing an LM's factual assertions back to specific training examples --- a task we call \emph{fact tracing}.
Training data attribution methods (TDA) are the main literature concerned with linking predictions back to specific training examples (known as ``proponents''). Influence functions \cite{hampel1974influence, koh2017understanding} and TracIn \cite{pruthi2020estimating} are among the first methods to do this for neural networks, by estimating the marginal effect of a training example on the loss of a test-time example. However, most work on TDA has focused on classification and regression tasks that do not necessarily involve fine-grained factual information \cite{Han2020ExplainingBB, hara2019data}.

Several obstacles have limited research on fact tracing for large, pre-trained LMs. First, since pre-training corpora are very large, it has not been clear how to obtain ground truth labels regarding which pre-training example was truly responsible for an LM's prediction. 
Second, TDA methods have traditionally been computationally prohibitive. 
In this paper, we present one of the first computationally tractable studies of fact tracing for LMs. To do so, we construct:
\begin{enumerate}[label=(\arabic*)]
    \item Two specially designed evaluation datasets, \textsc{FTRACE-TREx} and \textsc{FTRACE-Synth}, which contain unambiguous ground-truth information about the origin of specific facts.
    \item A tractable procedure for evaluating fact-tracing methods on large-scale LMs.
\end{enumerate}

\paragraph{Obtaining Ground Truth Proponents}
To establish (1) ground truth data for fact tracing, we propose a new recipe, which we call ``novel fact injection''. First, suppose that we can identify a set of ``facts'' that the pre-trained LM does \emph{not} know --- we call these ``novel facts''. We can convert each novel fact into an LM training example, and then fine-tune the LM on these extra examples until it memorizes the novel facts (i.e. ``injecting'' them into the LM). With a few caveats, we now know that the LM must have learned these facts from our newly injected examples. We also know which examples are responsible for teaching each fact, since we constructed each example from a particular fact. Hence, we now have ground-truth ``proponents'' for every novel fact, and can evaluate any TDA method on its ability to identify these proponents -- i.e. to retrieve the true proponents out of a large set of training examples. 

We implement this recipe using the TREx dataset \cite{elsahar2018t} as our source of novel facts. TREx is a large text corpus where each sentence has been comprehensively annotated with the facts that it expresses, in the form of relational knowledge tuples. To identify novel facts present in TREx, we filter for knowledge tuples that the pre-trained LM did not already know, as tested using masked LM prompting. The sentences in TREx expressing these tuples are then ``injected'' via fine-tuning and labeled as proponents. We call this setup \textbf{\textsc{FTRACE-TREx}}.

There are two caveats for the above setup. First, we must be careful about how we define what an LM ``knows''. For example, if an LM generates a particular assertion with 10\% probability, does this count as ``knowing'' or not? Second, some facts can be indirectly inferred from other facts. For example, suppose we want to know how an LM learned that Barack Obama was born in Hawaii. It could learn this from a literal mention of the fact: ``Obama was born in Hawaii'', or indirectly infer it from ``Obama was born in Honolulu''. TREx dataset mostly identifies literal proponents (the former), but not indirect proponents (the latter).

To address these two issues, we introduce an additional, more controlled setup, \textbf{\textsc{FTRACE-Synth}}, featuring synthetically generated novel facts that could not have possibly been known by the pre-trained LM, and which also have no correlation with any existing facts -- making indirect inferences impossible.

\paragraph{Mitigating Computational Cost}
To mitigate (2) the high computational cost of most TDA methods, we propose a simple reranking setup that is commonly used in information retrieval (IR) experiments. Rather than running a TDA method over all training examples, we run it only over a small subset of ``candidate'' examples that is guaranteed to include the ground truth proponents as well as some ``distractor'' examples that are not true proponents. In this way, a TDA method always has the opportunity to identify the true proponents while still facing challenging distractors, which enables us to differentiate the performance of multiple methods.

\paragraph{Key Results} Having developed data and quantitative evaluation methods for fact tracing, we use them to 
evaluate two popular families of TDA methods: gradient-based methods \citep[such as][]{pruthi2020estimating}, and embedding-based methods \cite{rajani2020explaining}.
As a reference point, we also compare these TDA methods against a simple baseline: BM25 \cite{robertson1995okapi, lv2011lower}, a standard IR technique that simply selects proponents by retrieving training examples that have high lexical overlap with the query. 

To measure the full potential of existing TDA methods, we optimize over key design choices including checkpoint selection, layer selection, embedding normalization and more. Furthermore, we significantly improve TracIn by introducing a simple but novel variant that accounts for training optimizer momentum \citep{shazeer2018adafactor}.
Even with these improvements, we find that much headroom remains, as all TDA methods still under-perform BM25 on the \textsc{FTRACE-TREx} dataset. This does not imply that BM25 is optimal for the task, but rather demonstrates clear ways in which TDA methods could improve.
On our more controlled \textsc{FTRACE-Synth}, we observe that TDA methods are significantly more competitive, especially when we introduce lexical variation in the way facts are expressed. We conclude that significant headroom remains for TDA methods to successfully address the new task of fact tracing.

\section{Retrieval Methods}
We begin with a formal description of the different TDA methods we study in this paper: \emph{gradient-based} methods \cite{koh2017understanding, pruthi2020estimating} and \emph{embedding-based} methods \cite{rajani2020explaining}.
To contextualize the performance of these two families of approaches, we also describe a widely used information retrieval baseline, BM25, which uses surface lexical similarity and thus tells us how effectively we can perform fact tracing without even having to access a model.

\subsection{Gradient-based Attribution}
Influence functions \cite{hampel1974influence, koh2017understanding} provide one of the first and best-known attribution methods. Given a training example $z = (x, y)$ and a test example $z_{\textrm{query}} = (x_{\textrm{query}}, y_{\textrm{query}})$, influence functions seek to estimate the change in the loss on $z_{\textrm{query}}$ given an $\epsilon$ increase in the weight of a particular training example $z$ at training time. Computing the influence of a training example $z$ involves first estimating the change in the optimal parameters $\hat{\theta}$, given that the example $z$ is up-weighted by $\epsilon$ in the training objective, then calculating how much the loss on $z_{\textrm{query}}$ changes w.r.t. the parameter change. The resulting influence score for convex loss functions is shown to be:
\begin{align}
\label{eq:influence_functions}
&\mathcal{I}(z, z_{\textrm{query}}) = \nonumber \\ &\quad -\nabla_{\theta} L\left(z_{\text {query}}, \hat{\theta}\right)^{\top} H_{\hat{\theta}}^{-1} \nabla_{\theta} L\left(z, \hat{\theta}\right)
\end{align}
where $\nabla_{\theta} L(z, \theta)$ denotes the gradient of the loss function on example $z$ evaluated at model parameters $\theta$, and $H_{\hat{\theta}}$ denotes the Hessian of the training objective evaluated at the final converged model parameters, $\hat{\theta}$ (see \citet{koh2017understanding} for the derivation).
In this form, influence functions can be roughly viewed as the weighted dot product of the gradients for $z_\textrm{query}$ and $z$, where the weight is the inverse Hessian of the training objective at $\hat{\theta}$. Due to the complexity of inverse Hessian calculation, the naive computational complexity is $\mathcal{O}(np^2 + p^3)$ ($n$ is dataset size, $p$ is parameter size). Even after the sampling approximations proposed in \citet{koh2017understanding}, the cost is still too high to directly apply influence functions for fact tracing.\footnote{\citet{schioppa2021scaling} propose more tractable approximations for Hessian based influence, but the memory requirement of the proposed method is still infeasible without projecting gradients into lower dimensions. Please refer to \cite{basu2021influence} for additional shortcomings.}

Therefore, we turn to a more recent TDA method that has demonstrated both better tractability and strong empirical results: TracIn \cite{pruthi2020estimating}, which seeks to estimate influence by asking a credit-assignment question rather than a counterfactual perturbation question. During training, when we take a gradient step on training example $z$ (input, output) at time $t$, we ask how much the loss changes on test example $z_{\textrm{query}}$. TracIn employs a first-order Taylor approximation to answer this question, yielding the following estimate, which is simply the dot product of gradients at a particular step $t$:
\begin{equation}\label{eq:dottracin}
\mathcal{I_\textrm{t}}(z, z_{\textrm{query}}) = \nabla_{\theta} L\left(z_{\text {query}}, \theta_{\textrm{t}}\right)^{\top}  \nabla_{\theta} L\left(z, \theta_{\textrm{t}}\right)
\end{equation}
If we have taken $K$ gradient steps on the training example, this yields the total influence:
\begin{align}\label{eq:tracinsum}
&\mathcal{I}(z, z_{\textrm{query}}) = \nonumber \\ &\qquad \sum_{k=1}^K \nabla_{\theta} L\left(z_{\text {query}}, \theta_{\textrm{t}(k)}\right)^{\top}  \nabla_{\theta} L\left(z, \theta_{\textrm{t}(k)}\right)
\end{align}
where $\textrm{t}(k)$ denotes the training step at which we took the $k^{th}$ gradient step on training example $z$.

The sum over time steps is generally approximated by using some fixed set of training checkpoints, which need not coincide with the actual steps where $z$ was visited. 
A known issue is that gradient similarity may be dominated by outlier training examples with large gradients. A simple fix proposed in previous work \cite{barshan2020relatif, han2021influence} is to unit-normalize the gradients, effectively replacing the dot product in \cref{eq:dottracin} with cosine similarity:
\begin{align}
&\mathcal{I}(z, z_{\textrm{query}}) = \nonumber \\ &\qquad \sum_{k=1}^K \frac{\nabla_{\theta} L\left(z_{\text {query}}, \theta_{\textrm{t}(k)}\right)^{\top}\nabla_{\theta} L\left(z, \theta_{\textrm{t}(k)}\right)}{\|\nabla_{\theta}L\left(z_{\text {query}}, \theta_{\textrm{t}(k)}\right)\|\|\nabla_{\theta}L\left(z, \theta_{\textrm{t}(k)}\right)\|}
\label{eq:cosgrad}
\end{align}
We hereafter refer to $\mathcal{I}$ in \cref{eq:cosgrad} as \influencemethod. %
\subsection{Embedding-based Attribution}
Hidden representations of neural networks are known to embed high-level features that are often useful for similarity search. While not as theoretically justified, prior work \cite{rajani2019explain} has found that such representations can outperform gradient-based methods in model interpretability tasks.
Following prior work, we extract the intermediate layer outputs of a Transformer language model, and average over decoding time-steps to obtain a single vector representation for any example. In our experiments, we consider representations at different layers of the Transformers, as well as their concatenations. Similar to the case of gradient-based methods, the association between a training example and a model prediction is defined by a cosine product: 
\begin{align}
&\mathcal{I}(z, z_{\textrm{query}}) = \nonumber \\ &\qquad \frac{\mathit{LM}_{\textrm{inter.}}(z)^{\top} ® \mathit{LM}_{\textrm{inter.}}(z_{\textrm{query}})}{\|\mathit{LM}_{\textrm{inter.}}(z)^{\top}\|\|\mathit{LM}_{\textrm{inter.}}(z_{\textrm{query}})\|}
\label{eq:embed}
\end{align}
where $\mathit{LM}_\text{inter.}$ denotes a layer's output in $\mathit{LM}$. We refer to $\mathcal{I}$ in \cref{eq:embed} as \embeddingmethod.

\subsection{Baseline: BM25}
In the previous sections, we used attribution methods to define a model-specific similarity function between examples.
But it is also possible to identify facts in a model-agnostic way: In the classic IR literature, word-overlap based methods have been shown to be both simple and effective.

Among these approaches, BM25 \cite{robertson1995okapi, lv2011lower}, the best performing variant, has been consistently used as a baseline for information retrieval benchmarks \cite{thakur2021beir} . When using BM25, we consider an example as a bag of words consist of the input and the output words. The score is proportional to token overlap between the query and the candidate, inversely weighted with the frequency of such tokens, and the importance of weights regulated by hyperparameters (\cref{app:impdetail}).

\section{Fact Tracing Datasets}
We propose two datasets to measure fact tracing approaches: \textsc{FTRACE-TREx}, a natural language dataset with real facts derived from the TREx dataset, and \textsc{FTRACE-Synth}, a synthetic dataset with novel facts using made-up entities and relations. For each dataset, we define an \textbf{attribution set} containing all LM training examples that might be considered proponents and a \textbf{query set} containing test examples, each annotated with their ground truth proponents from the attribution set. The examples in these sets consist of masked inputs, outputs and the fact annotations.
\subsection{FTRACE-TREx}
\label{sec:dataset}
\begin{figure}
    \centering
    \includegraphics[width=\linewidth]{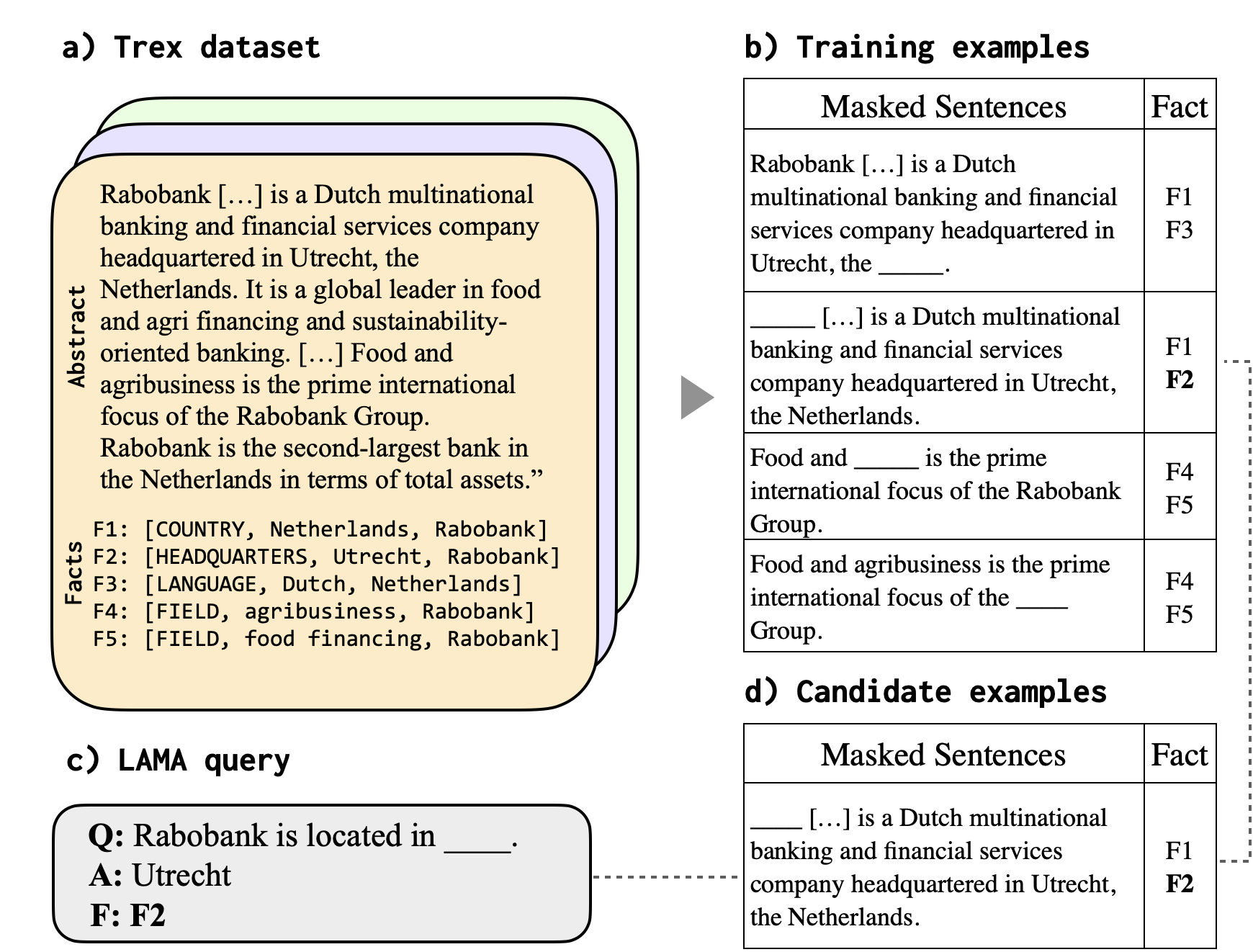}
    \caption{Dataset Creation: From the original TREx \cite{elsahar2018t} data, we construct masked sentences and annotate their facts by using provided fact annotations. We identify proponents by matching the TREx sentences expressing the same fact. (The outputs of the masked examples are omitted in the figure.)}
    \label{fig:dataset}
\end{figure}

\input{tables/dataset}
We create an attibution set using TREx \cite{elsahar2018t} and query set using LAMA \cite{petroni2019language} datasets. 
TREx consists of DBPedia \cite{brummer2016dbpedia} abstracts, $a_i \in A$. Each abstract contains a set of sentences, $s_j = a_{ij}$, and each sentence is associated with a set of facts, $F(s_{j})$. For each fact $f \in F(s_j)$, TREx annotates the exact positions where the subject and object respectively appear in the sentence $s_j$.

We wish to convert these sentences into training examples that can teach a language model about the facts stated within them. 
To do so, we construct cloze-style language modeling examples as in masked language modeling~\cite{Devlin2019BERTPO} or span corruption~\cite{raffel2019exploring}. In particular, for each fact $f$ in a sentence $s$, we mask out either the subject or the object, and train the model to predict it. The two resulting examples $\text{mask}_\text{sub}(s, f)$ and $\text{mask}_\text{obj}(s, f)$ are marked as ``proponents'' of the fact, as shown in \cref{fig:dataset}.

The LAMA dataset is anchored to the same fact tuples used by TREx. For each fact tuple, LAMA provides a template-generated sentence expressing the fact. Similar to TREx, we convert this sentence into a cloze-style example by either masking out the subject or object.
Hence, we now have two sets of examples (TREx and LAMA) that express the same facts. We treat the TREx examples as our attribution set and the LAMA examples as our test set. Since we wish to trace influence from LAMA back to TREx, we sometimes refer to LAMA examples as ``queries'' and TREx examples as ``retrieval candidates.''
For any LAMA example, we define the ground-truth proponents as simply the TREx examples that express the same fact. 

One ambiguity remains regarding ground truth in TREx sentences that express multiple facts. Suppose a TREx sentence expresses facts $f_1$ and $f_2$, and we generate cloze examples for both $f_1$ and $f_2$. The example $\text{mask}_\text{sub}(s, f_1)$ is clearly a proponent of $f_1$, but it is perhaps also a proponent of $f_2$, since the text supporting $f_2$ is still present after masking. Ultimately, we care about whether attribution methods can retrieve the right \emph{sentence} from the attribution set, not a particular \emph{masking} of that sentence. In our evaluations (described next), we evaluate a method's ability to retrieve at the \emph{sentence level}, with the score of a sentence defined as the max score over all maskings of that sentence.

In total (\cref{tab:datastats}), we match approximately \textbf{448k} TREx sentences with \textbf{31k} LAMA queries. On average, each TREx example expresses three facts, and each LAMA example has 83 proponents (including different maskings of the same sentence).
\subsection{FTRACE-Synth}
In a dataset with real facts, two factors can negatively impact TDA methods  compared to baselines such as BM25: First, many of the facts in \textsc{FTRACE-TREx} may already be known by a pre-trained LM.\footnote{Even if the answer is not ranked first among model outputs, the correct prediction may be “close to the surface” on these examples, and the contribution of fine-tuning may be small, even if the predictions flip.} In such cases, the LM will not learn the fact from TREx, and TDA methods should not be expected to identify examples in TREx as proponents. We refer to this as the ``saturation'' problem, since the model's performance already saturated on the fact before fine-tuning, leaving no signal for TDA methods to detect. Second, real corpora like TREx and LAMA have lexical overlap between query and attribution examples (overlapping surface forms; see \cref{sec:dataset}) which can favor counting-based methods like BM25.

To better evaluate TDA methods in isolation, we create a synthetic dataset, \textsc{FTRACE-Synth} with facts that are guaranteed to be novel. First, we create random entities with a total number comparable to \textsc{TREx}. Then, we randomly relate those entities with each other using the same set of relations from the TREx dataset.
\paragraph{Entities}
Our entity list consists of 5,000 synthetic entities each uniquely identified by a number. To reduce the lexical overlap between examples in the dataset, we use 4 surface forms per entity -- 2 forms with Arabic numerals, 2 forms with Roman numerals. For example, the fourth entity appears with the following surface forms:    
["4-entity", "entity-4", "IV-entity", "entity-IV"].

\paragraph{Relations}
The dataset includes a set of 37 relations (\cref{app:synthrels}) borrowed directly from TREx. Additionally, we paraphrase each relation to create diversity and to reduce the lexical overlap between attribution and query examples.

\paragraph{Attribution Set}
Each example in the attribution corpus expresses two facts to parallel the multi-fact nature of TREx examples.
\begin{center}
\fbox { \parbox { 0.95\linewidth} {
\textbf{Input:} entity-MMCLXXIV is the official language of \rule{1cm}{0.1mm}$_1$ \textcolor{red}{,} CMXCVII-entity is the writing place of \rule{1cm}{0.1mm}$_2$ \\[.25em]
\textbf{Output:} 1:3082-entity, 2:entity-MMMCCC}}
\end{center}
The attribution corpus includes 50,000 individual facts. By masking different entities as well as combining different facts, we can generate 3,190,000 masked examples for the attribution corpus.
\paragraph{Query Set}
Similar to LAMA, each example in the query corpus queries a single fact expressed as a masked example, for example:
\begin{center}
\fbox { \parbox { 0.95\linewidth} {
\textbf{Input:} entity-3300 was written in \rule{1cm}{0.1mm}.\\[.25em]
\textbf{Output:} entity-CMXCVII
}}
\end{center}
We generate 5,000 such facts by assigning random relations between different entities, with two surface forms for each, resulting in 10,000 examples. As a result, each fact in this query set has 62 proponents in the attribution corpus, and every entity appears in 10 relations on average.

\section{Experimental Setup}\label{sec:expsetup}
Our experiments aim to answer the questions of (1) whether TDA methods can be used as effective fact tracing tools (compared to simple IR baselines), (2) which configurations make them most effective (exploring many variations), and (3) analyzing the weaknesses of \influencemethod, in particular its sensitivity to \emph{when} the knowledge is learned (the aforementioned ``saturation'' hypothesis). 

\subsection{Reranking Evaluation}\label{sec:ranking}
Ideally, an attribution method would score a given test query against every training example, and we can sort all examples by their influence score. %
This would enable evaluation with standard IR methods like
recall@10 and mean reciprocal rank (MRR) $\frac{1}{|Q|} \sum_{q \in Q} \frac{1}{\operatorname{rank}_{q}}$,
where $\operatorname{rank}_{q}$ is the rank of the first true proponent for the query, and $Q$ denotes the candidate set.
However, most attribution methods are computationally intractable for scoring all training sentences in large datasets. Although we can reduce the complexity of some of these methods through the use of random projections \cite{pruthi2020estimating}, such lossy approximations would render our results less conclusive, as it would be unclear whether an outcome is due to the intrinsic quality of a method or the quality of the projection.

Therefore, to achieve computational tractability while avoiding such confounds, we propose a simple reranking setup: instead of scoring all examples, we can score a carefully selected subset that still enables meaningful comparisons. We call this the ``candidate set''. It is the union of four sets:
\begin{enumerate}
    \item all true proponents for a query: $\mathcal{P}(z_\textrm{query})$,
    \item the top-100 retrievals from BM25: $\operatorname{BM25}(z_\textrm{query})$,
    \item 100 random examples that share the same target $y$ as the query:  $\mathcal{D}_{y}=\{(x,y)\text{ s.t. } y=y_{\textrm{query}}\}$, and
    \item 100 randomly sampled examples: $\mathcal{D}_{\textrm{random}}$,
\end{enumerate} 
with random samples fixed across all evaluations.
Note that MRR on this particular candidate set is an \textbf{upper-bound} on the MRR over the full attribution set. Because it includes all proponents but fewer distractors, rank is guaranteed to be closer to 1 in the MRR definition. Also, including $\mathcal{P}(z_\textrm{query})$ is necessary to ensure that the model has the opportunity to retrieve all proponents.
$\mathrm{BM}25(z_\text{query})$ ensures that we have ``distractors'' with high lexical overlap, and  $\mathcal{D}_{y}$ is included because we observed that TDA methods have a tendency to retrieve examples with the same output as the query.

Our candidate set includes all top retrievals from BM25, so the results for BM25 are exact. When combined with the fact that reranking MRRs always upper-bound full retrieval MRRs, our setup guarantees that 
any method that underperforms BM25 on reranking also underperforms on full retrieval.
\paragraph{Slicing examples}

The gradient-based methods require careful treatment when considering models that go through two separate stages: pre-training and fine-tuning. For example, if a model has already obtained zero loss on an example at the start of fine-tuning, then the gradient will be near-zero throughout fine-tuning, and computing influence using only fine-tuning checkpoints will yield an uninformative influence score for any query.
We refer to this problem as ``saturation.'' To mitiagate saturation, we evaluate TDA methods on a subset of queries we label \textbf{Finetune-learned (FL)},  where the model failed before fine-tuning (the answer is not in top-3 beam-search predictions), but succeeded afterward (the answer is ranked first in the beam). We referred to this set as ``novel facts'' in Section~\ref{sec:intro}.\footnote{In \cref{app:submetrics}, we present additional results for \textbf{Pretrain-learned (PL)} examples, which went from failing to successful during pre-training rather than fine-tuning.}

\subsection{Model}\label{sec:model}
We use MT5-base, a commonly used encoder-decoder language model \cite{xue2020mt5} to evaluate the aforementioned neural TDA methods. MT5 was pre-trained on the MC4 corpus, which includes all of Wikipedia, and therefore was exposed to the knowledge expressed in \textsc{FTRACE-TREx}. 
The pre-trained MT5 model achieves 24.3\% top-3 accuracy when predicting answers to the \textsc{TREx} queries.
Fine-tuning MT5 on our \textsc{FTRACE-TREx} training set increases accuracy to 47.42\%, suggesting that there are still many facts MT5 did not know after pre-training. For \textsc{FTRACE-Synth}, the pre-trained model gets 0 accuracy as expected, and the fine-tuned model obtains 81\%\footnote{We accept an answer if any of the surface forms of the correct entity is the output.}.

To evaluate \influencemethod{}, we approximate \cref{eq:tracinsum} by choosing three checkpoints that are uniformly spaced out in terms of their training loss (specifically, inverse perplexity), to ensure that we cover significant parts of training while favoring regions with greater loss reduction. Note that we use pre-training checkpoints when evaluating the pre-trained model, and fine-tuning checkpoints when evaluating the fine-tuned model; see \cref{app:impdetail} for details.
We calculate the gradient w.r.t the average negative likelihood of the true output token sequence. To evaluate embedding-based fact tracing, we use representations from the final checkpoint of the model. %

For both gradient and embedding-based methods, we present the best layer combination among the different concatenations of layers studied in (\cref{sec:layers}).

\section{Results}\label{sec:Discussion}
\begin{table}[t]
    \centering
        \resizebox{\linewidth}{!}{%
        \footnotesize
    \begin{tabular}{@{}lllll@{}}
    \toprule
         \bf Methods & \multicolumn{2}{c}{\bf MRR} &  \multicolumn{2}{c}{\bf Recall@10} \\
           \midrule
        Random-Target & \multicolumn{2}{c}{14.50 \stderr{0.95}} & \multicolumn{2}{c}{10.32 \stderr{1.54}}\\
        BM25 & \multicolumn{2}{c}{\bf 77.55 \stderr{1.50}} & \multicolumn{2}{c}{	60.89 \stderr{2.31}	} \\
        \midrule
            & \multicolumn{1}{c}{\bf Finetuned} &  \multicolumn{1}{c}{\bf Pretrained} & \multicolumn{1}{c}{\bf Finetuned} &  \multicolumn{1}{c}{\bf Pretrained}\\
                 \midrule
        \influencemethod &48.56 \stderr{4.40}&	62.38 \stderr{1.99}	&56.02 \stderr{0.67}&	57.54 \stderr{1.25}\\
        \embeddingmethod &	64.29 \stderr{1.32}&	60.59 \stderr{1.13}&	57.89 \stderr{1.38}&	54.91 \stderr{0.32}  \\
        \influencemethod + \embeddingmethod & 58.52 \stderr{3.83}&	67.66 \stderr{0.22}&	61.72 \stderr{0.08}&	61.59 \stderr{1.35}\\
        \bottomrule
    \end{tabular}
     }
    \caption{Top Level Results: Best scores for each method and model on the \textbf{Finetune-learned} slice of \textsc{\bf FTRACE-TREx}. We present average  sentence-level retrieval results over 3 random selections of 200 queries. We found that BM25 performs best in MRR outperforming neural methods.  \cref{tab:retrievalwsub1} shows detailed MRRs on predicate, subject, and object level of candidate examples.}
    \label{tab:retrieval}
\end{table}

\subsection{Top-level comparisons}
\label{sec:mainresults}

In \cref{tab:retrieval}, we present a top-level comparison of the three main methods discussed (gradient-based, embedding-based, and BM25). Hyperparameters for all methods have been optimized. As we discuss in subsequent sections, TDA hyperparameters have a significant effect on performance.

We optimized \influencemethod by rescaling gradients with Adafactor accumulators \cite{shazeer2018adafactor}, applying unit-normalization to the gradients (see \cref{tab:variation}) and selecting the best layer configuration (\cref{sec:layers}).  To sanity check that TDAs are doing more than matching the query's output label, we compare to a  \textsc{Random-Target} baseline that outputs a score of 1 for all training examples with the same output label. This baseline is indeed substantially worse than either method. 

Despite extensive optimization for \influencemethod and \embeddingmethod, however, we found that BM25 with no tuning still outperforms neural TDAs in MRR and Recall@10. \influencemethod slightly outperforms \embeddingmethod for pretrained model but significantly underforms \embeddingmethod for the finetuned model. When we ensemble \influencemethod and \embeddingmethod (by summing their influence scores) there are improvements on both metrics, demonstrating that their success is somewhat orthogonal. 
We provide example retrievals from all three models in \cref{app:samples}.

Surprisingly, pre-trained \influencemethod outperforms fine-tuned \influencemethod in this dataset, as we discuss more in \cref{sec:synthres}.

We do not seek to measure all benefits of attribution methods, but rather to assess one \textbf{expected} function (fact-tracing), as promised by their stated goal (tracing a model's prediction back to data). 
The fact that even the best TDA method obtains MRR of 67.66 and Recall@10 of 61.59 showcases the significant \emph{absolute} headroom that remains for attribution methods . BM25 results are only a little better, and are provided mainly as a reference point. Next, we analyze what choices contributed to the current status of TDA methods with a detailed exploration of hyperparameters.

\subsection{Which Transformer layers provide the most reliable attribution signal?}
\label{sec:layers}
\begin{figure*}[t]
    \centering    \includegraphics[width=\linewidth]{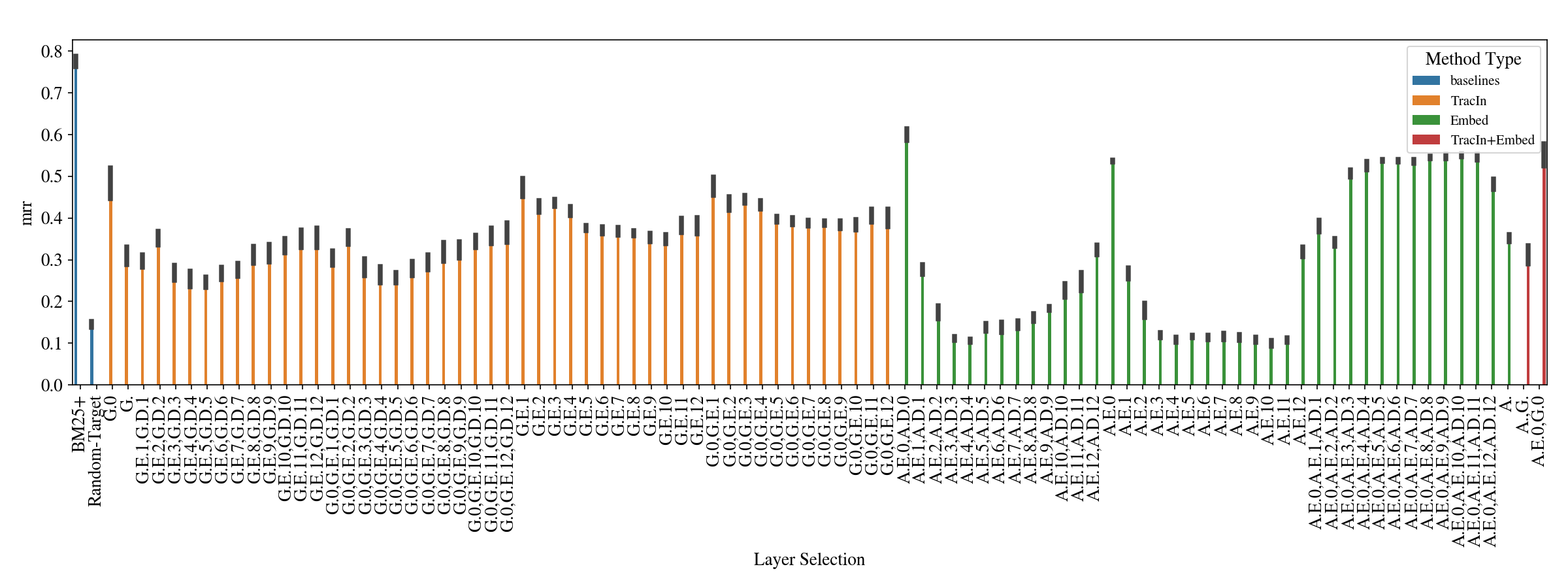}
    \vspace{-2em}
    \caption{Mean reciprocal rank for \textcolor{myorange}{\influencemethod} with different layers and and \textcolor{mygreen}{\embeddingmethod} from different intermediate layers: In \textbf{G.0}, gradient of embedding layer is used. In \textbf{G.} and \textbf{A.}, respectively, gradients and embeddings of all layers are used. \textbf{A.E.0} and \textbf{A.D.0} corresponds to embedding layer's output in the encoder and decoder part of the model respectively. Comma-separated labels denote ensembling by summing the scores of the corresponding layers. We report results for 3 random seeds (error bars with standard deviation) of 200 queries where queries learned between pretraining checkpoints. In neural methods, using only the embedding layer or its output performs the best, while underperforming the \textcolor{myblue}{baseline method} BM25.}
    \label{fig:layers}
\end{figure*}

Some layers of a language model may be specialized for operations that have no relation to factual information. For example, previous probing work \cite{tenney2019bert} shows the existence of layers that focus on syntax rather than on knowledge. The contribution of such layers to \influencemethod may introduce noise. In \cref{fig:layers}, we conduct an experiment where we sweep over various subsets of layers.

For \influencemethod, the best-performing layer is the embedding layer of the model --- this result, also observed in \citet{yeh2022first}, is surprising, as most prior work used only the last layer. In \embeddingmethod, the best performing layer is again the output of the embedding layer. These results suggest that much of the effectiveness of embedding-based methods derives from their models of lexical similarity. Conversely, for \influencemethod, the embedding layer also includes contextual information since the gradient signal propagates back through the entire model.

\paragraph{Additional Model Variants}
\begin{table}[t]
    \centering
    \resizebox{\linewidth}{!}{%
    \begin{tabular}{@{}lllll@{}}
    \toprule
         & \multicolumn{2}{c}{\bf MRR} &  \multicolumn{2}{c}{\bf Recall@10} \\
    \midrule
         \bf Change & \multicolumn{1}{c}{\bf Finetuned} &  \multicolumn{1}{c}{\bf Pretrained} & \multicolumn{1}{c}{\bf Finetuned} &  \multicolumn{1}{c}{\bf Pretrained}\\
    \midrule
       Adafactor $\xrightarrow{}$ no-Adafactor &  \textcolor{red}{--3.83 \stderr{ 4.81}} &	\textcolor{red}{--7.20 \stderr{ 2.25}}	& \textcolor{red}{--11.29 \stderr{ 2.05}}	& \textcolor{red}{--2.36 \stderr{ 1.63}} 	\\
        unit-norm $\xrightarrow{}$ no-norm & \textcolor{red}{--3.36 \stderr{ 4.89}} &	\textcolor{red}{--32.90 \stderr{ 2.13}} &	\textcolor{red}{--10.82 \stderr{ 2.24}}	& \textcolor{red}{--28.06 \stderr{ 1.46}} \\
        multi-ckpt  $\xrightarrow{}$ single-ckpt& \textcolor{blue}{+0.51 \stderr{6.42}}	&\textcolor{blue}{+0.60 \stderr{ 2.23}} & \textcolor{red}{--6.95 \stderr{ 4.72}}	& \textcolor{blue}{+5.44 \stderr{ 1.61}} \\
        no [eos]  $\xrightarrow{}$  [eos] & \textcolor{blue}{+5.50 \stderr{ 4.63}}&	\textcolor{red}{--24.77 \stderr{ 3.82}} &	\textcolor{blue}{+12.96 \stderr{1.59}} &	\textcolor{red}{--19.93 \stderr{ 3.49}} 	\\
        \bottomrule
    \end{tabular}
    }
    \caption{Our experiment with various configurations for best layer of the \influencemethod evaluted in Finetune-Learned set of \textsc{FTRACE-TREx}: For each change from the best configuration (the first row), we report the best result by optimizing free hyper-parameters. The normalization and the usage of the accumulator smoothing was effective in our top level \influencemethod results. We compare maximum scored checkpoint scores to our original multi-checkpoint results, we found that the best checkpoint performs slightly better than multi-heckpoint results in MRR.  The including the the end of sentence token in the target side hurts pretrained MT5 model since it is originally trained to predict multiple answer.
    \vspace{-.5em}
    }
    \label{tab:variation}
\end{table}
\cref{sec:mainresults} mentioned several design choices for \influencemethod. We performed a systematic evaluation of those choices. 
In \cref{tab:variation}, given the set of configurable options in the table, we set a given option to a particular value and then optimize remaining parameters. 

Unit-normalized gradients drastically outperform the dot product. Next, we considered the role of Adafactor during training.
The \influencemethod equation arises from considering parameter updates at a specific step. The true parameter updates were not raw gradients, but gradients that had been rescaled by Adafactor accumulators. Using these rescaled gradient for \influencemethod performs much better. Also, surprisingly, using the single best-performing checkpoint is sometimes better than using multiple checkpoints. We provide the individual checkpoint results in \cref{tab:singleckpt}.

\subsection{\textsc{FTRACE-Synth} and Saturation}\label{sec:synthres}

\begin{table}[t]
    \centering
        \resizebox{0.9\linewidth}{!}{%
        \footnotesize
    \begin{tabular}{@{}llll@{}}
    \toprule
         & \multicolumn{1}{c}{\bf MRR} &  \multicolumn{1}{c}{\bf Precision @10}&\multicolumn{1}{c}{\bf Recall@10} \\
    \midrule
        Random-Target &36.47  \stderr{2.84} &	30.43 \stderr{4.00}	& 2.45 \stderr{0.32}\\
        BM25 & 87.69 \stderr{1.71}&	52.02 \stderr{2.65}	& 4.20 \stderr{0.21}\\
        \midrule
        \influencemethod & 100.00 \stderr{0.00}	&  99.50 \stderr{0.14}	&   8.02 \stderr{0.01} \\
        \embeddingmethod & 99.58 \stderr{0.24} &	97.12 \stderr{0.53}	& 7.83 \stderr{0.04}   \\
        \influencemethod + \embeddingmethod & 100.00 \stderr{0.00}	& 98.07 \stderr{0.18} & 	7.91 \stderr{0.01} \\
        \bottomrule
    \end{tabular}
     }
    \caption{Synthetic Dataset Results: Best scores for fine-tuned model on the \textbf{Finetune-learned} slice of \textsc{ FTRACE-Synth}. On completely novel facts, the TracIn upperbound is higher than the other methods. Since we control the lexical overlap, BM25 underperforms neural methods. We present average  sentence-level retrieval results over 2 random selections of 200 queries. The upper-bound scores on neural methods are higher in the synthetic data than BM25 as we reduce the lexical overlap. The \influencemethod upperbound performs best in all the metrics.}
    \label{tab:synth}
\end{table}
As mentioned earlier, \influencemethod{} monitors the change in a model's performance on a test query over the course of training --- and therefore is likely to fail if a test query's loss is already zero (saturated) at the start of the training period monitored by \influencemethod{}. In addition, because the pre-trained model sees very similar sentences and information in the pre-training corpora, the influence could be distributed over multiple examples, such that the signal from each candidate is weak. These confounding factors may apply to \textsc{FTRACE-TREx}. Therefore, we also evaluate TDA methods on our synthetic dataset, \textsc{FTRACE-Synth}, which controls for all these issues. We fine-tune the same model on \textsc{FTRACE-Synth} and perform the same evaluation in \cref{tab:synth}. The results suggest that when the aforementioned factors are controlled, the reranking upper-bound for gradient-based TDA method is better than BM25 and slightly better than embedding-based TDA methods. This result verifies that TDA methods might have advantages over standard IR methods, despite falling short in a more realistic, applied scenario.

\section{Related work}
\label{sec:background}
\paragraph{Information Retrieval}
To define our fact tracing task, we employ standard concepts from the information retrieval (IR) literature: a retrieval + reranking setup, and standard retrieval metrics. However, while IR focuses on retrieving any document that satisfies a user's query, our benchmark specifically aims to identify examples that caused a particular model to make a particular prediction. This focus on model-specific causality distinguishes us from prior IR work \cite{thakur2021beir, nguyen2016ms}. Our evaluation setup should be \emph{easier} than generic IR benchmarks because we are only evaluating on predictions we know the LM gets right.

\paragraph{Language Models as Retrievers}
Language models have been successfully used in numerous IR applications. \citet{karpukhin2020dense} use language model embeddings to warm-start neural retrievers for knowledge-intensive tasks. \citet{guu2020realm} and \citet{ lewis2020retrieval} show that language modeling and information retrieval can be jointly learned in a manner that benefits both tasks. Our work uses TDA-based retrieval methods to help users understand the behavior of the LMs themselves.

\paragraph{Attribution Methods}
Recent work has tried to explain neural model behavior in many different ways: (1) attributing a prediction back to specific features in the input \cite{Simonyan2013DeepIC, Sundararajan2017AxiomaticAF, Han2020ExplainingBB}, (2) attributing to specific model parameters \cite{dai2021knowledge,mitchell2021fast}, (3) probing for competence at linguistic sub-tasks \cite{tenney2019bert}, and finally (4) attributing back to training examples \cite{pruthi2020estimating, koh2017understanding}.

However, work in the last category \cite{Han2020ExplainingBB, guo2020fastif, zhang-etal-2021-sample} has been limited, mainly focusing on classification and regression tasks that do not involve questions about factuality or world knowledge. Consequently, these methods have primarily been used as a data cleaning technique, leaving the question of fact tracing unexplored \cite{Han2020ExplainingBB, hara2019data}.

\section{Conclusion}
\label{sec:conclusion}
We introduced a new dataset and benchmark for \emph{fact tracing}: the task of tracing language models' assertions back to the training examples that provided evidence for those predictions. We evaluated \emph{gradient}-based and \emph{embedding}-based TDA methods and found that they perform worse than a standard IR baseline (BM25) even in settings that favor TDA methods. We investigated the effects of layer selection, model checkpoints and fine-tuning. Our ablative analysis suggests that saturation is an important factor affecting the performance of current methods. Much is needed to improve these methods before they can be reliably used for fact tracing. We hope that this benchmark will enable future research on fact tracing, by mitigating computational challenges and establishing a principled ground truth.

\section*{Acknowledgments}
We would like to thank Zhuyun Dai, Keith B. Hall, Ji Ma for their helpful discussions and feedbacks.

\section*{Limitations and Impact Statement}
Our experiments focus on a single representative language model, MT5-base; it is possible that our findings about the effectiveness of attribution methods for fact tracing would differ substantially when applied to language models with very different architectures or trained on different datasets. Because of the candidate set construction scheme described in \cref{sec:ranking}, these results only upper-bound the performance of evaluated methods, and it is also possible that they are even less effective than reported here. The ground truth labels in \textsc{FTRACE-TREx} extracted from TREx where the fact annotations are \emph{semi-automatically} annotated, can have labeling errors.

The FTRACE dataset includes content from Wikipedia, some of which has not been vetted for factual accuracy. It is possible that by redistributing this content we will propagate misinformation. We plan to mitigate this harm with a datasheet that explicitly communicates FTRACE's role as an evaluation tool, and not as a reliable source of information. Apart from the dataset, we anticipate no ethical issues associated with the techniques described in this publication.

\bibliography{references}
\bibliographystyle{acl_natbib}
\onecolumn
\appendix
\textbf{\huge{Appendix}}
\vspace{0.75cm}

In this appendix, we will provide implementation details and additional results for the experiments.

\section{Implementation Details}\label{app:impdetail}
\paragraph{BM25} 
We use the following BM25 formula:
\begin{equation}
\mathcal{I}(z,  z_{\textrm{query}})=\sum_{t \in z_{\textrm{query}}}  \log \left(\frac{N+1}{N_{t}}\right) \nonumber \\ \times \left(\frac{\left(k_{1}+1\right) \cdot f(z, t)}{k_{1} \cdot\left((1-b)+b \cdot\left(\frac{L(z)}{L_\text{avg}}\right)\right)+f(z, t)}+ 1\right)
\end{equation}

where, f(z, t) is the overlap count, $N$ is the number of training examples, $L(z)$ is the length of the example, and $L_\text{avg}$ is the average example length. $k_1$ and $b$ are hyperparameters tha reweights the importance of the other terms in the formula. \citet{robertson1995okapi} provides the intuition behind this definition of relatedness.

We use a publicly available BM25+\cite{lv2011lower} implementation written in python and released under \url{https://pypi.org/project/rank-bm25/}. We tokenize queries and retrieval examples by space and we remove masked tokens. We did not optimize any of the default hyper parameters.
\paragraph{MT5 Model} We use intermediate checkpoints of MT5 model \footnote{https://github.com/google-research/multilingual-t5} (12 layers transformer with 580M parameters). We convert these checkpoints to Pytorch by using HugginFace's T5 converter. We use the tokenizer provided. In our dataset\cref{sec:dataset}, we use \texttt{extra\_id\_0} for the mask token compatible with pretraining corpus of MT5..  
\paragraph{\influencemethod} We calculate gradients by using Pytorch without batching examples and by using average negative likelihood over output sequence. We store each individual parameter's gradient (blocks of transformer) in a dictionary structure. Given a query and a retrieval example, we calculate scores \cref{eq:cosgrad} for each parameter seperately that means we locally normalize each parameters' gradient in \cref{eq:cosgrad}. Then, to calculate a layer's or full model's score, we score individual scores corresponding to parameters in that layer. This enable us to sweep over different combination of layers as in \cref{fig:layers} without rerunning the model. 

Pretrained MT5 model is trained until 80k gradient steps. We use checkpoints at 5100, 10200, 15300 steps. We fine-tune MT5 model on additional 60k gradient steps on TREx dataset. Then, we use checkpoints at 5000, 10000, 30000 steps.

We paralelize over checkpoints when calculating \cref{eq:cosgrad}. For each query, we spend approximately 15 minutes by using VOLTA V100 32 GB GPUs to get scores for all the retrieval examples in the ranking set (\cref{sec:ranking}))
\paragraph{\embeddingmethod} Transformer model's forward pass can be expressed as following pseudo code:
\begin{align}
\label{eq:transformer}
\begin{split}
     \textrm{enc}_0 &= \textrm{Embedding}(x) \\
     \textrm{enc}_i &= \textrm{Encoder}_{i}(\textrm{enc}_{i-1})  i=1..N \\ 
      \textrm{dec}_0 &= \textrm{Embedding}(y) \\
     \textrm{dec}_i &= \textrm{Decoder}_{i}(y, \textrm{enc}_{N})  i=1..N \\
    \mathcal{L} &= \textrm{NLL}(W_{\textrm{proj}}\textrm{dec}_N,y_{\textrm{query}})
\end{split}
\end{align} 

We use $\textrm{enc}_i$ and $\textrm{dec}_i$, and reduce (average) them over time-steps in input and outputs side respectively.
\newpage
\section{Synthetic Data Relation Templates}\label{app:synthrels}
Below are the relation templates we use in the dataset. ``0 and ``1" are the slots for the entities. Paraphrases are delimited by ``|" sign. Left paraphrase is the original surface for in the \textsc{FTRACE-TREx} dataset, right one is the additional paraphrase paraphrase.
\lstset{basicstyle=\ttfamily\footnotesize,breaklines=true}
\begin{lstlisting}
{0} was born in {1} | {0}'s birth place is {1}   
{0} died in {1} | {0} passed away in {1}   
{0} is a subclass of {1} | {1} is superclass of {0}   
The official language of {0} is {1} | {1} is the official language of {0}   
{0} plays in {1} position | {1} is the play position of {0}   
{0} was awarded the {1} | {1} given to {0}   
{0} was originally aired on {1} | {1} is the first streamer of {0}   
{0} was educated at the University of {1} | {0} studied in University of {1}    
{0} shares border with {1} | {0} and {1} are neighbours   
{0} is named after {1} | {1} was inspirational for the naming of {0}   
The original language of {0} is {1} | {1} is the original language of {0}   
{0} plays with {1} | {0} plays along with {1}   
{0} is a member of {1} | {1} accepted {0} as a member   
{0} works in the field of {1} | {1} is the work field of {0}   
{1} participated in the {0} | {1} was a participant of {0}   
{0} is a {1} by profession | {0}'s profession is {1}   
{0} consists of {1} | {0} includes {1}   
{0} is a member of the {1} political party | {0}'s political party was {1}   
{0} maintains diplomatic relations with {1} | {0}'s diplomacy with {1}   
{0} is produced by {1} | {1} produced {0}   
{0} is a citizen of {1} | {0}'s home country is {1}   
{0} was written in {1} | {1} is the writing place of {0}   
{0} is located in {1} | {0} placed in {1}   
{0} is developed by {1} | {1} developed {0}   
{0} is the capital of {1} | the capital of {1} is {0}   
{0} works for {1} | {0} works at {1}   
{0} plays {1} music | {0} perform {1} music   
{0} has the position of {1} | {0}'s position is {1}   
{0} is represented by music label {1} | music label {1} represents {0}   
{0} used to work in {1} | {1} is ex-workplace of {0}   
{0} is affiliated with the {1} religion | {0} believes in {1} religion   
{0} is owned by {1} | {1} owned {0}v
The native language of {0} is {1} | {1} is the native language of {0}   
{0} and {1} are twin cities | {0} is twin city of {1}   
{0} is a legal term in {1} | {0} is a legal definition in {1}   
The headquarter of {0} is in {1} | {0}'s headquarter in {1}   
{0} was founded in {1} | {0} was established in {1}   
\end{lstlisting}
\section{Additional Results and Samples}\label{app:samples}
\subsection{Individual Checkpoints}
\begin{table}[h!]
\centering
\begin{tabular}{lllll}
\toprule
&\multicolumn{2}{c}{MRR} & \multicolumn{2}{c}{Recall@10}  \\
\midrule
&FT &  PT &    FT  &  PT\\
\midrule
Multi &    48.56\stderr{      4.40} &      62.38\stderr{      1.99} &      56.02\stderr{0.67} &      57.54\stderr{1.25} \\
\midrule
Ckpt1 &     49.07\stderr{      4.67} &      54.77\stderr{      1.26} &      49.07\stderr{4.67} &      54.77\stderr{1.26} \\
Ckpt2 &      47.30\stderr{      2.88} &      62.98\stderr{      1.01} &      47.30\stderr{      2.88} &      62.98\stderr{1.01} \\
Ckpt3 &     48.69\stderr{      5.19} &      60.29\stderr{      3.34} &      48.69\stderr{5.19} &      60.29\stderr{3.34} \\
\bottomrule
\end{tabular}
    \caption{\influencemethod results for individual checkpoints on Finetune-Learned set.}
    \label{tab:singleckpt}
\end{table}
\subsection{MRR Results with Submetrics}\label{app:submetrics}
\paragraph{MRR on Finetune-Learned Subsets of FTRACE-TREx}
We provide submetrics for (\textbf{Finetuned-learned (FL)} ) set.
\begin{table*}[!h]
    \centering
        \caption{MRR Results with submetrics in fine-tuned learned set. (see \cref{tab:retrieval})}
        \resizebox{\linewidth}{!}{%
        \footnotesize
    \begin{tabular}{lllllllll}
    \toprule
       & \multicolumn{2}{c}{Sentence (Table 2) } &  \multicolumn{2}{c}{Predicate} &  \multicolumn{2}{c}{Subject} &  \multicolumn{2}{c}{Object}  \\
    \midrule
     &\multicolumn{1}{c}{FT}&\multicolumn{1}{c}{PT}&\multicolumn{1}{c}{FT}&\multicolumn{1}{c}{PT}&\multicolumn{1}{c}{FT}&\multicolumn{1}{c}{PT}&\multicolumn{1}{c}{FT}&\multicolumn{1}{c}{PT}\\
    \midrule
        Random-Target & 14.50\stderr{0.95}	&14.50\stderr{0.95} &14.71\stderr{0.89}&	14.71\stderr{0.89}& 98.14\stderr{0.72}	&98.14\stderr{0.72} & 63.56\stderr{2.53}	&63.56\stderr{2.53}\\
        BM25 & 77.55\stderr{1.50} &	77.55\stderr{1.50} & 79.26\stderr{2.82}	&79.26\stderr{2.82} &88.25\stderr{1.80}	&88.25\stderr{1.80} 	& 85.71\stderr{1.22}&	85.71\stderr{1.22} \\
        \midrule
        \influencemethod & 48.56\stderr{4.40} &	62.38\stderr{1.99}& 49.16\stderr{4.76}	&63.98\stderr{0.98} &	99.53\stderr{0.43}	& 86.49\stderr{1.22} & 88.74\stderr{1.59}	&74.99\stderr{3.61}\\
        \embeddingmethod & 64.29\stderr{1.32} &		60.59\stderr{1.13} & 66.25\stderr{1.82}	& 63.00\stderr{1.73} & 94.09\stderr{0.77}	& 81.79\stderr{1.33} & 80.45\stderr{0.99}	& 74.03\stderr{2.27} \\
        \influencemethod + \embeddingmethod & 58.52\stderr{3.83} &	67.66\stderr{0.22} & 59.24\stderr{3.88} &	69.49\stderr{0.92} & 97.92\stderr{0.50}	& 82.03\stderr{1.61} & 71.94\stderr{2.44} &	79.15\stderr{1.55}
 \\
        \bottomrule
    \end{tabular}
    }
    \label{tab:retrievalwsub1}
\end{table*}
\paragraph{MRR on Pretrained-Learned Subsets of FTRACE-TREx}
We present additional results for (\textbf{Pretrain-learned (PL)} ) examples where the model failed before the a checkpoint of pre-training, but changed during pre-training. We found that the average number of proponents in the PL set is 2.5x that of the FL set (since we expect that frequently mentioned facts will be learned first). These results suggest that it’s difficult to control for when facts were learned without affecting the other statistics, and that direct comparisons between model performance on the PL and FL datasets may not be informative. 
\begin{table*}[!h]
    \centering
        \caption{MRR Results with submetrics in pretrained- learned set. (see \cref{tab:retrieval})}
        \resizebox{\linewidth}{!}{%
        \footnotesize
    \begin{tabular}{lllllllll}
    \toprule
       & \multicolumn{2}{c}{Sentence (Table 2) } &  \multicolumn{2}{c}{Predicate} &  \multicolumn{2}{c}{Subject} &  \multicolumn{2}{c}{Object}  \\
    \midrule
     &\multicolumn{1}{c}{FT}&\multicolumn{1}{c}{PT}&\multicolumn{1}{c}{FT}&\multicolumn{1}{c}{PT}&\multicolumn{1}{c}{FT}&\multicolumn{1}{c}{PT}&\multicolumn{1}{c}{FT}&\multicolumn{1}{c}{PT}\\
    \midrule
        Random-Target & 15.83\stderr{1.42}	 &15.83\stderr{1.42} & 15.62\stderr{1.29}&	15.62\stderr{1.29} & 98.36\stderr{0.79}	& 98.36\stderr{0.79} & 61.88\stderr{1.99}		&61.88\stderr{1.99} \\
        BM25 & 77.62\stderr{3.24} &	77.62\stderr{3.24} & 77.20\stderr{3.56} &	77.20\stderr{3.56} &91.24\stderr{0.83}& 	91.24\stderr{0.83}&  88.07\stderr{1.39}	& 88.07\stderr{1.39}\\
        \midrule
        \influencemethod & 64.18\stderr{2.62} &	54.45\stderr{1.96} & 63.94\stderr{1.80}	&56.01\stderr{1.95} & 99.17\stderr{0.72}	& 89.82\stderr{2.21}& 88.07\stderr{2.00} &	81.44\stderr{1.71} \\
        \embeddingmethod & 51.21\stderr{2.43}	& 50.42\stderr{2.43} &51.02\stderr{2.55}		&50.30\stderr{2.64} & 96.52\stderr{1.75}	& 84.21\stderr{3.03} & 79.18\stderr{1.05}	&79.00\stderr{0.82}\\
        \influencemethod + \embeddingmethod & 65.91\stderr{2.88}	& 55.40\stderr{1.98} 	&66.04\stderr{2.60}	& 57.09\stderr{2.24} & 97.95\stderr{0.25}& 	86.21\stderr{2.16} & 84.09\stderr{2.00}&	82.01\stderr{1.78}
 \\
        \bottomrule
    \end{tabular}
    }
    \label{tab:retrievalwsub2}
\end{table*}

\subsection{Precision-Recall plots for FTRACE-TREx}
We present accompanying precision and recall results for \cref{fig:layers}.

\includegraphics[width=0.95\linewidth]{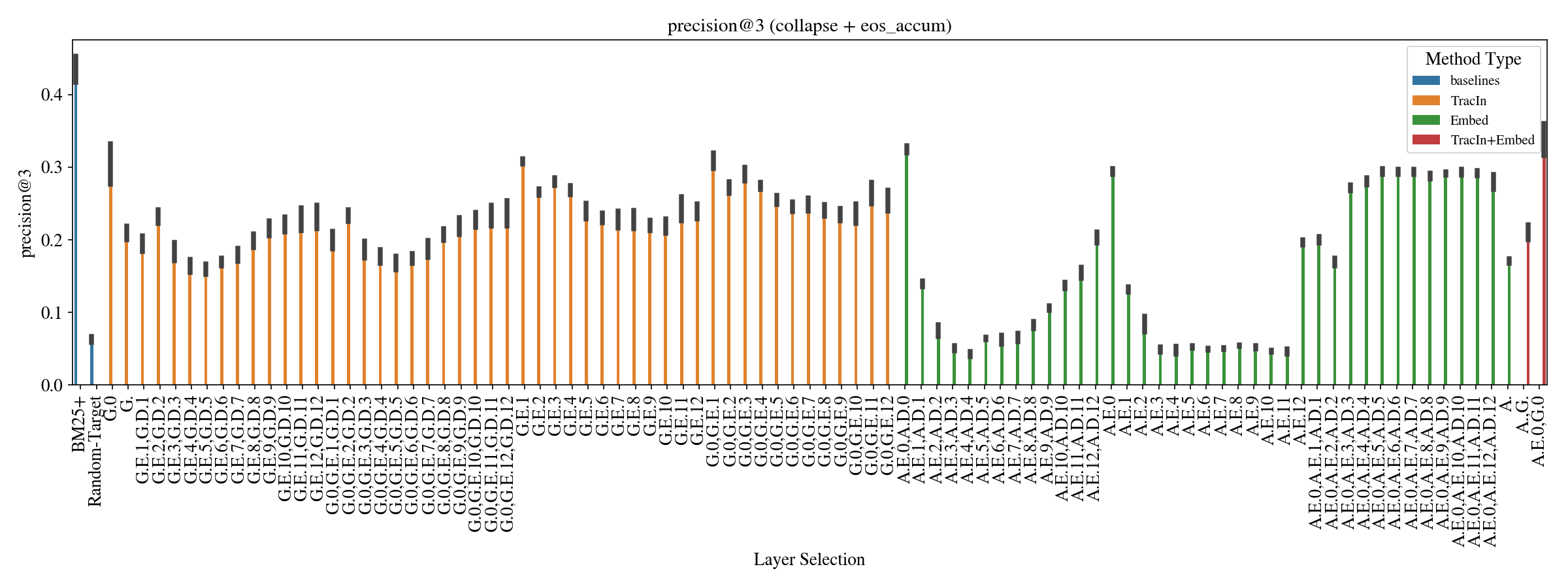}

\includegraphics[width=0.95\linewidth]{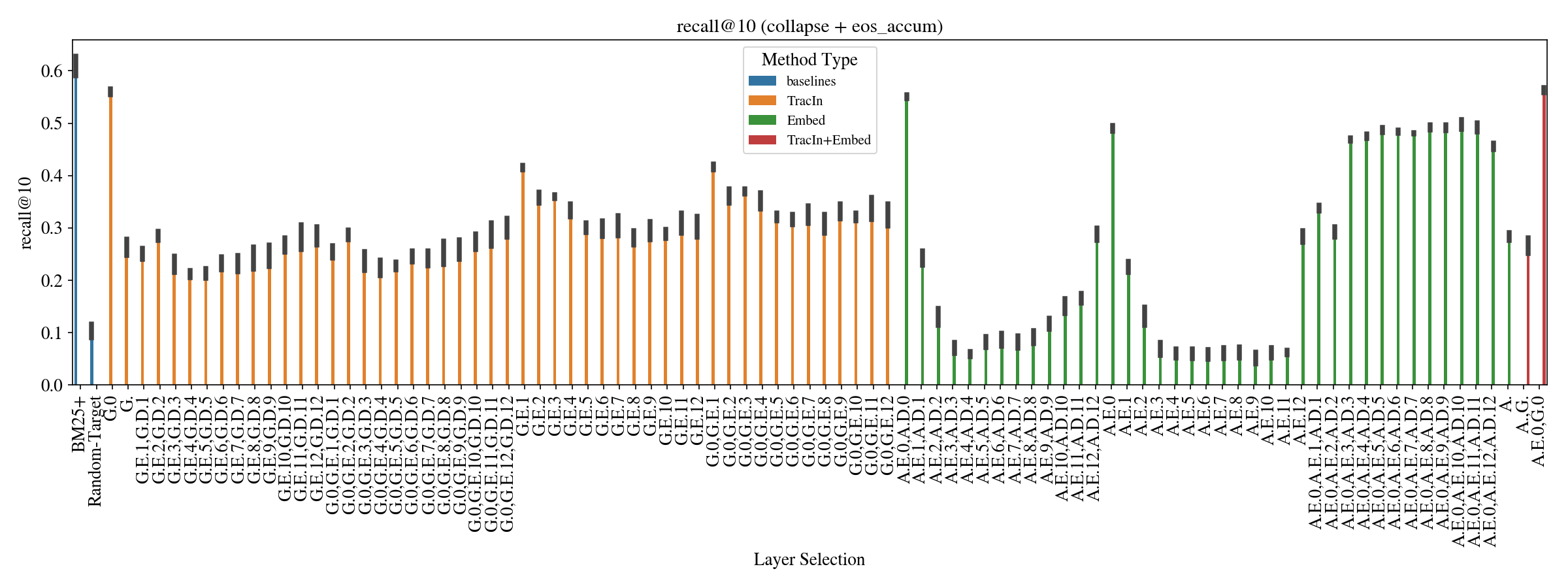}

\subsection{Samples for FTRACE-TREx}\label{app:samplestrex}
Here, we provide example top-3 retrievals from TDAs for the \textsc{FTRACE-TREx} dataset. Long examples are truncated for display purposes. We provide label (whether the retrieved example includes the fact) next to the output of the retrieved example.
\begin{table}[h!]
    \centering
         \resizebox{0.95\linewidth}{!}{%
\begin{tabularx}{\textwidth}{{X|X|X}}
\toprule
Embed &TracIn & BM25 \\  
\midrule
\textbf{Q}: In late 2005, the \rule{1cm}{0.1mm}$_1$ broadcast a full series of Star Spell, again presented by Eamonn Holmes but Mishal Husain took over from Nina as word pronou...\newline\textbf{A}: BBC \textbf{True} & \textbf{Q}: In late 2005, the \rule{1cm}{0.1mm}$_1$ broadcast a full series of Star Spell, again presented by Eamonn Holmes but Mishal Husain took over from Nina as word pronou...\newline\textbf{A}: BBC \textbf{True} & \textbf{Q}: In late 2005, the \rule{1cm}{0.1mm}$_1$ broadcast a full series of Star Spell, again presented by Eamonn Holmes but Mishal Husain took over from Nina as word pronou...\newline\textbf{A}: BBC \textbf{True} \\ \hline
\textbf{Q}: The Vicar of Dibley is a \rule{1cm}{0.1mm}$_1$ television sitcom created by Richard Curtis and written for actress Dawn French by Curtis and Paul Mayhew-Archer, wit...\newline\textbf{A}: BBC \textbf{False} & \textbf{Q}: Tasneem Zehra Husain (also spelled as Tasneem Zehra Hussain), is a Pakistani \rule{1cm}{0.1mm}$_1$ and an Assistant Professor of Physics at the Lahore University of...\newline\textbf{A}: theoretical physicist \textbf{False} & \textbf{Q}: In late 2005, the BBC broadcast a full series of Star Spell, again presented by Eamonn Holmes but \rule{1cm}{0.1mm}$_1$ took over from Nina as word pronouncer.\newline\textbf{A}: Mishal Husain \textbf{True} \\ \hline
\textbf{Q}: Honigberg also recorded Homage to Rostropovich (1927–2007), a CD of solo cello works written for the legendary cellist; Frédéric Chopin's complete wor...\newline\textbf{A}: piano \textbf{False} & \textbf{Q}: Abdul Aziz Bin Dato Haji Husain was born 18 July 1950 in Kuching, Sarawak, \rule{1cm}{0.1mm}$_1$.\newline\textbf{A}: Malaysia \textbf{False} & \textbf{Q}: He now works for the BBC, presenting on the BBC News channel and \rule{1cm}{0.1mm}$_1$.\newline\textbf{A}: BBC One \textbf{False} \\ 
\bottomrule
\end{tabularx}
}
    \caption{Mishal Husain works for \rule{1cm}{0.1mm}$_1$. (A: BBC)}
    \label{tab:ex1trex}
\end{table}

\begin{table}[h!]
    \centering
         \resizebox{0.95\linewidth}{!}{%
\begin{tabularx}{\textwidth}{{X|X|X}}
\toprule
Embed &TracIn & BM25 \\  
\midrule
\textbf{Q}: Clara Ellaline Hope Leighton (sometimes Clare Veronica Hope Leighton) (12 April 1898 - 4 November 1989) was an \rule{1cm}{0.1mm}$_1$/American artist, writer and ill...\newline\textbf{A}: English \textbf{False} & \textbf{Q}: He was educated in \rule{1cm}{0.1mm}$_1$ and at the Quaker Leighton Park School.\newline\textbf{A}: London \textbf{False} & \textbf{Q}: The \rule{1cm}{0.1mm}$_1$ Kenneth Leighton (1929–1988) also wrote a Fantasia Contrappuntistica ("Homage to Bach", Op.24) for piano, which won the first prize at the...\newline\textbf{A}: composer \textbf{True} \\ \hline
\textbf{Q}: Lillianne Brown Leighton (May 17, 1874 – March 19, 1956), known professionally as Lillian Leighton, was an \rule{1cm}{0.1mm}$_1$ silent film actress.\newline\textbf{A}: American \textbf{False} & \textbf{Q}: Kenneth \rule{1cm}{0.1mm}$_1$ Bray (May 26, 1895 – January 9, 1953) was an Episcopal priest, teacher, sportsman and coach.\newline\textbf{A}: Augustine \textbf{False} & \textbf{Q}: The composer \rule{1cm}{0.1mm}$_1$ (1929–1988) also wrote a Fantasia Contrappuntistica ("Homage to Bach", Op.24) for piano, which won the first prize at the Bolzano...\newline\textbf{A}: Kenneth Leighton \textbf{True} \\ \hline
\textbf{Q}: The composer Kenneth Leighton (1929–1988) also wrote a Fantasia Contrappuntistica ("Homage to Bach", Op.24) for \rule{1cm}{0.1mm}$_1$, which won the first prize at ...\newline\textbf{A}: piano \textbf{True} & \textbf{Q}: Leighton Road Evangelical Church is a nonconformist independent evangelical church located on the Gainsborough estate, \rule{1cm}{0.1mm}$_1$ in the English county o...\newline\textbf{A}: Ipswich \textbf{False} & \textbf{Q}: The composer Kenneth Leighton (1929–1988) also wrote a Fantasia Contrappuntistica ("Homage to Bach", Op.24) for \rule{1cm}{0.1mm}$_1$, which won the first prize at ...\newline\textbf{A}: piano \textbf{True} \\               
\bottomrule
\end{tabularx}
}
    \caption{Query: Kenneth Leighton plays \rule{1cm}{0.1mm}$_1$. (A: piano)}
    \label{tab:ex2trex}
\end{table}

\subsection{Samples for FTRACE-Synth}\label{app:samplessynth}
Now, we provide the retrieved examples for \textsc{FTRACE-Synth} version of our dataset.
\begin{table}[h!]
    \centering
         \resizebox{0.95\linewidth}{!}{%
\begin{tabularx}{\textwidth}{{X|X|X}}
\toprule
Embed &TracIn & BM25 \\  
\midrule
\textbf{Q}: \rule{1cm}{0.1mm}$_1$ given to 3692-entity,entity-2686 was awarded the \rule{1cm}{0.1mm}$_2$\newline\textbf{A}: 1:entity-1138, 2:entity-MMMMDCLIII \textbf{True} & \textbf{Q}: \rule{1cm}{0.1mm}$_1$ given to entity-MMDCLXXXVI,\rule{1cm}{0.1mm}$_2$ used to work in CXVI-entity\newline\textbf{A}: 1:entity-MMMMDCLIII, 2:entity-1650 \textbf{True} & \textbf{Q}: \rule{1cm}{0.1mm}$_1$ given to 3692-entity,\rule{1cm}{0.1mm}$_2$ was awarded the entity-MMMMDCLIII\newline\textbf{A}: 1:entity-1138, 2:entity-2686 \textbf{True} \\ \hline
\textbf{Q}: entity-1138 given to \rule{1cm}{0.1mm}$_1$,entity-2686 was awarded the \rule{1cm}{0.1mm}$_2$\newline\textbf{A}: 1:3692-entity, 2:entity-MMMMDCLIII \textbf{True} & \textbf{Q}: entity-CCCII given to \rule{1cm}{0.1mm}$_1$,MMMDLVI-entity given to \rule{1cm}{0.1mm}$_2$\newline\textbf{A}: 1:entity-MMMMDCLIII, 2:entity-MDCCCLXXXVII \textbf{False} & \textbf{Q}: entity-CCCII given to \rule{1cm}{0.1mm}$_1$,\rule{1cm}{0.1mm}$_2$ given to entity-MDCCCLXXXVII\newline\textbf{A}: 1:entity-MMMMDCLIII, 2:MMMDLVI-entity \textbf{False} \\ \hline
\textbf{Q}: \rule{1cm}{0.1mm}$_1$ given to 2686-entity,\rule{1cm}{0.1mm}$_2$ plays in entity-2658 position\newline\textbf{A}: 1:MMMMDCLIII-entity, 2:MMMMCCLXIX-entity \textbf{True} & \textbf{Q}: \rule{1cm}{0.1mm}$_1$ shares border with entity-DCCXXVII,entity-302 given to \rule{1cm}{0.1mm}$_2$\newline\textbf{A}: 1:entity-MMMMCMXCVIII, 2:entity-MMMMDCLIII \textbf{False} & \textbf{Q}: entity-CCCII given to \rule{1cm}{0.1mm}$_1$,MMMDLVI-entity given to \rule{1cm}{0.1mm}$_2$\newline\textbf{A}: 1:entity-MMMMDCLIII, 2:entity-MDCCCLXXXVII \textbf{False} \\ 
\bottomrule
\end{tabularx}
}
    \caption{\rule{1cm}{0.1mm}$_1$ given to entity-2686. (A: entity-MMMMDCLIII)}
    \label{tab:ex1:sytnh}
\end{table}

\begin{table}[h!]
    \centering
         \resizebox{0.95\linewidth}{!}{%
\begin{tabularx}{\textwidth}{{X|X|X}}
\toprule
Embed &TracIn & BM25 \\  
\midrule
\textbf{Q}: entity-MMMDLXXVI's diplomacy with \rule{1cm}{0.1mm}$_1$,entity-3193's birth place is \rule{1cm}{0.1mm}$_2$\newline\textbf{A}: 1:MMCDLX-entity, 2:entity-5 \textbf{True} & \textbf{Q}: 3132-entity maintains diplomatic relations with \rule{1cm}{0.1mm}$_1$,\rule{1cm}{0.1mm}$_2$ was awarded the entity-3701\newline\textbf{A}: 1:entity-3468, 2:entity-4097 \textbf{False} & \textbf{Q}: MMMDLXXVI-entity's profession is \rule{1cm}{0.1mm}$_1$,entity-CCLXI's diplomacy with \rule{1cm}{0.1mm}$_2$\newline\textbf{A}: 1:MXXX-entity, 2: 506-entity \textbf{False} \\ \hline
\textbf{Q}: MMMDLXXVI-entity's diplomacy with \rule{1cm}{0.1mm}$_1$,\rule{1cm}{0.1mm}$_2$ given to 2897-entity\newline\textbf{A}: 1:entity-MMCDLX, 2:MCMLIII-entity \textbf{True} & \textbf{Q}: The original language of MMCCLXXIX-entity is \rule{1cm}{0.1mm}$_1$,3552-entity shares border with \rule{1cm}{0.1mm}$_2$\newline\textbf{A}: 1:MMCDLX-entity, 2:MMMMDLXV-entity \textbf{False} & \textbf{Q}: MMMDLXXVI-entity's diplomacy with \rule{1cm}{0.1mm}$_1$,MCMLIII-entity given to \rule{1cm}{0.1mm}$_2$\newline\textbf{A}: 1:entity-MMCDLX, 2: 2897-entity \textbf{True} \\ \hline 
\textbf{Q}: MMMDLXXVI-entity's diplomacy with \rule{1cm}{0.1mm}$_1$,MCMLIII-entity given to \rule{1cm}{0.1mm}$_2$\newline\textbf{A}: 1:entity-MMCDLX, 2: 2897-entity \textbf{True} & \textbf{Q}: The official language of CMXCVII-entity is \rule{1cm}{0.1mm}$_1$,\rule{1cm}{0.1mm}$_2$ died in MMCDLX-entity\newline\textbf{A}: 1:3215-entity, 2: 710-entity \textbf{False} & \textbf{Q}: MMMDLXXVI-entity's diplomacy with \rule{1cm}{0.1mm}$_1$,\rule{1cm}{0.1mm}$_2$ given to 2897-entity\newline\textbf{A}: 1:entity-MMCDLX, 2:MCMLIII-entity \textbf{True} \\ 
\bottomrule
\end{tabularx}
}
    \caption{MMMDLXXVI-entity's diplomacy with \rule{1cm}{0.1mm}$_1$.  (A: MMCDLX-entity)}
    \label{tab:ex2:sytnh}
\end{table}
\end{document}

%% file: tables/dataset.tex
\begin{table}[t]
    \centering
     \resizebox{0.95\linewidth}{!}{%
    \begin{tabular}{@{}lllll@{}}
    \toprule
           &  \multicolumn{2}{c}{\bf \textsc{FTRACE-TREx}} &  \multicolumn{2}{c}{\bf \textsc{FTRACE-Synth}}  \\
          \bf Statistics & \bf Attribution & \bf Query & \bf Attribution &\bf Query \\
    \midrule
         Length & 1,560,453 & 31,479 & 3,190,000 &10,000 \\
         Unique Facts & 552,381 & 31,479 &50,000   & 5,000\\
         Avg. \#proponents & -- & 83 & -- & 62 \\
        Facts per example & 8.28  & 1 & 2 & 1\\
         Unique Predicates & 488 & 41&   37 & 37 \\
         Unique Objects & 49,166 & 2,266& 5,000 & 5,000  \\
         Unique Subjects & 310,197 & 29,464& 5,000 & 5,000  \\

         \bottomrule
    \end{tabular}
    }
    \caption{\textsc{FTRACE-TREx}: We extract 1M masked examples from TREx \cite{elsahar2018t}, and match them with 27k queries from  LAMA \cite{petroni2019language}  to construct our fact tracing benchmark. \textsc{FTRACE-Synth}: To evaluate influence methods on completely novel facts, we propose a synthetic benchmark consists of made-up entities and relations. Refer to \cref{app:samplestrex} and \cref{app:samplessynth} for examples.}
    \label{tab:datastats}
\end{table}